\documentclass[journal]{IEEEtran}
\usepackage{times}
\usepackage{epsfig}
\usepackage{amsmath}
\usepackage{amssymb}
\usepackage{multirow}
\usepackage{graphicx}
\usepackage{subfigure}
\usepackage{eso-pic}
\usepackage{cite}
\usepackage{color}
\usepackage{adjustbox}
\usepackage{url}

\def\tblue#1{\textcolor[rgb]{0,0,1}{#1}} 
\def\tred#1{\textcolor[rgb]{1,0,0}{#1}}  

\ifCLASSINFOpdf
\else
\fi
\begin{document}

\title{Mind the Gap: Enlarging the Domain Gap in Open Set Domain Adaptation}

\author{Dongliang Chang\inst{1}\and
Aneeshan Sain\inst{2} \and
Zhanyu Ma\inst{1} \and
Yi-Zhe Song\inst{2} \and
Jun Guo \inst{1}}

\author{Dongliang~Chang,~Aneeshan~Sain,~Zhanyu~Ma,~Yi-Zhe~Song, and~Jun~Guo 

\thanks{D. Chang, Z. Ma, J. Guo are with the Pattern Recognition and Intelligent
System Laboratory, School of Artificial Intelligence, Beijing University of Posts and Telecommunications, Beijing 100876, China (e-mail: mazhanyu@bupt.edu.cn).}

\thanks{A Sain and Y.-Z. Song  are with the  Centre for Vision, Speech and Signal Processing, University of Surrey, London, United Kingdom.}}

\maketitle

\begin{abstract}
Unsupervised domain adaptation aims to leverage labeled data from a source domain to learn a classifier for an unlabeled target domain. Among its many variants, open set domain adaptation (OSDA) is perhaps the most challenging, as it further assumes the presence of  unknown classes in the target domain. In this paper, we study OSDA with a particular focus on enriching its ability to traverse across larger domain gaps. Firstly, we show that existing state-of-the-art methods suffer a considerable performance drop in the presence of larger domain gaps, especially on a new dataset (PACS) that we re-purposed for OSDA. We then propose a novel framework to specifically address the larger domain gaps. The key insight lies with how we exploit the mutually beneficial information between two networks; (a) to separate samples of known and unknown classes, (b) to maximize the domain confusion between source and target domain without the influence of unknown samples. It follows that (a) and (b) will mutually supervise each other and alternate until convergence. Extensive experiments are conducted on Office-$31$, Office-Home, and PACS datasets, demonstrating the superiority of our method in comparison to other state-of-the-arts. Code  available  at~\url{https://github.com/dongliangchang/Mutual-to-Separate/}

\end{abstract}


%
\IEEEpeerreviewmaketitle

\section{Introduction}\label{Intro}

\begin{figure}[t] 
  \centering 
  \subfigure[]{
    \label{fig:OSDA:1a}
    \includegraphics[width=0.8\linewidth]{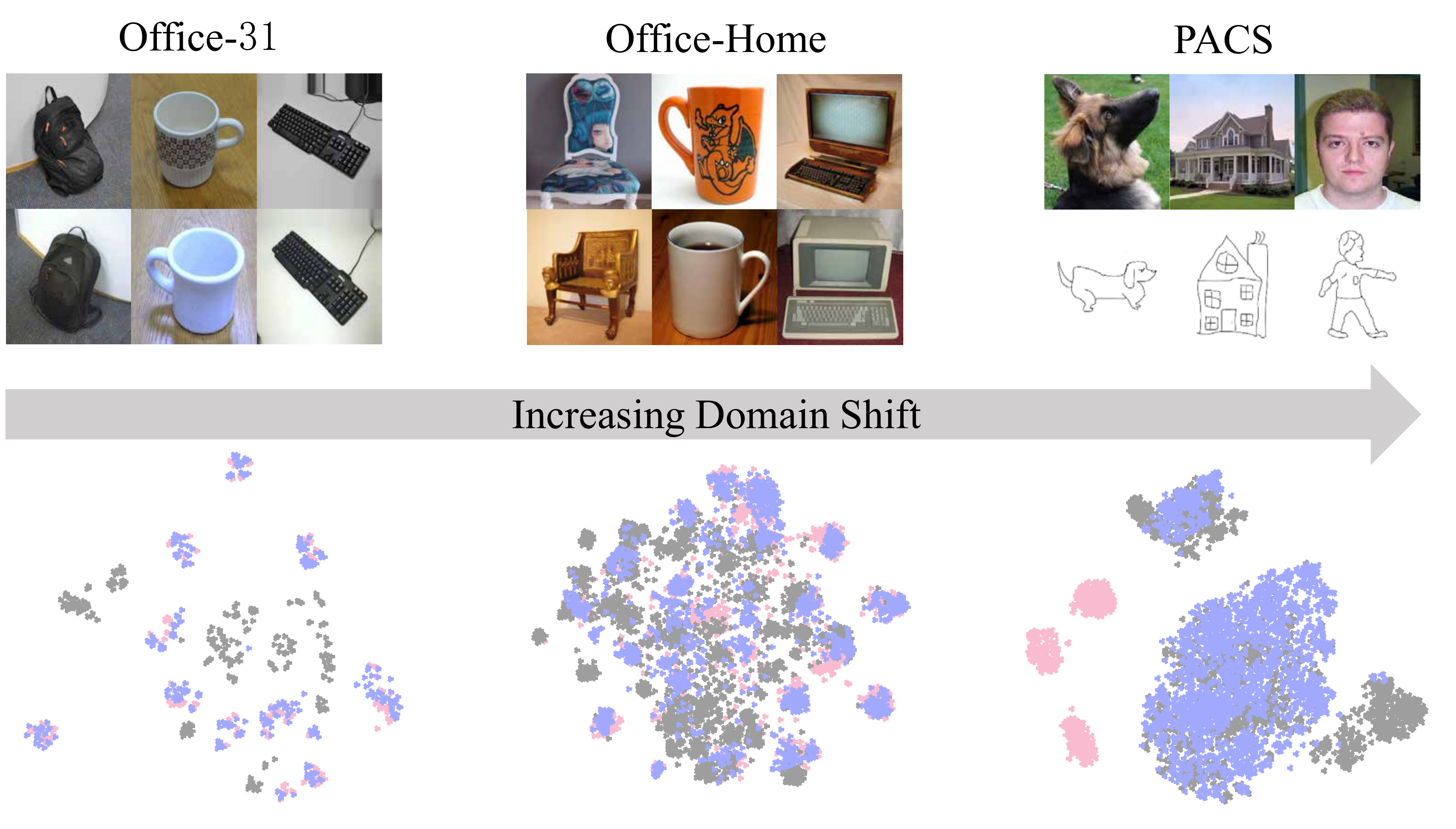}}
  \quad 
  \subfigure[]{
    \label{fig:OSDA:1b}
    \includegraphics[width=0.8\linewidth]{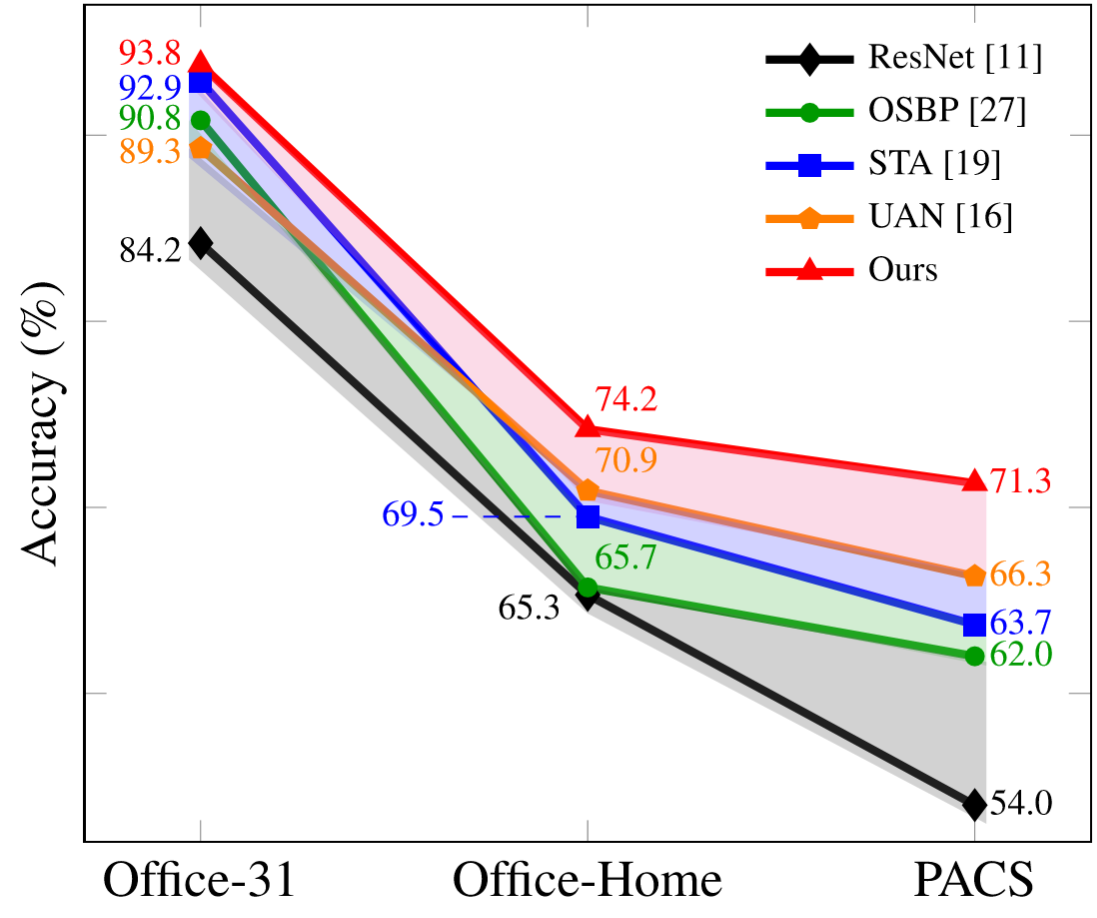}} 
  \caption{(a) Example images of three open set domain adaptation datasets with visualization of features extracted by ResNet$50$ on different tasks using t-SNE embeddings~(selecting the most difficult domain adaptation task as the source and the target in each dataset). Pink points are source features, blue and grey points refer to target features of known and unknown classes respectively. 
  (b) Classification accuracy(\%) on three open set domain adaptation tasks (ResNet-$50$). Please see Section~\ref{ALLExperiments} for details. As evident from the shaded region, our method exhibits the lowest relative performance drop with increasing domain shift.~\emph{Best viewed in color and zoomed in.}}
  \label{fig:OSDA}
\end{figure}

Great strides have been witnessed on the practical application of computer vision in recent years~\cite{yu2016sketch,ganin2014unsupervised,Kaichao2019Universal}. The remarkable efficacy of deep learning, however, relies heavily on the availability of an abundance of annotated data, which can be tedious and often impractical to collect. 
This has led to a recent surge of research asking an alternative question -- can we borrow off-the-shelf datasets from an existing source domains to benefit the training of a new target domain?
In an ideal scenario, where the source domain data shares the same underlying distribution with the target, this would have been straightforward. However, that is rarely the case -- domain gaps naturally exist as a result of different illumination conditions (light vs. dark), capturing devices (webcam vs. DSLR), styles (photo vs. painting), and abstraction levels (photo vs. sketch). The key challenge presented to all domain adaptation algorithms is therefore how best to address the domain gap.

\begin{figure*}[t]
\begin{center}
   \includegraphics[width=1.0\linewidth]{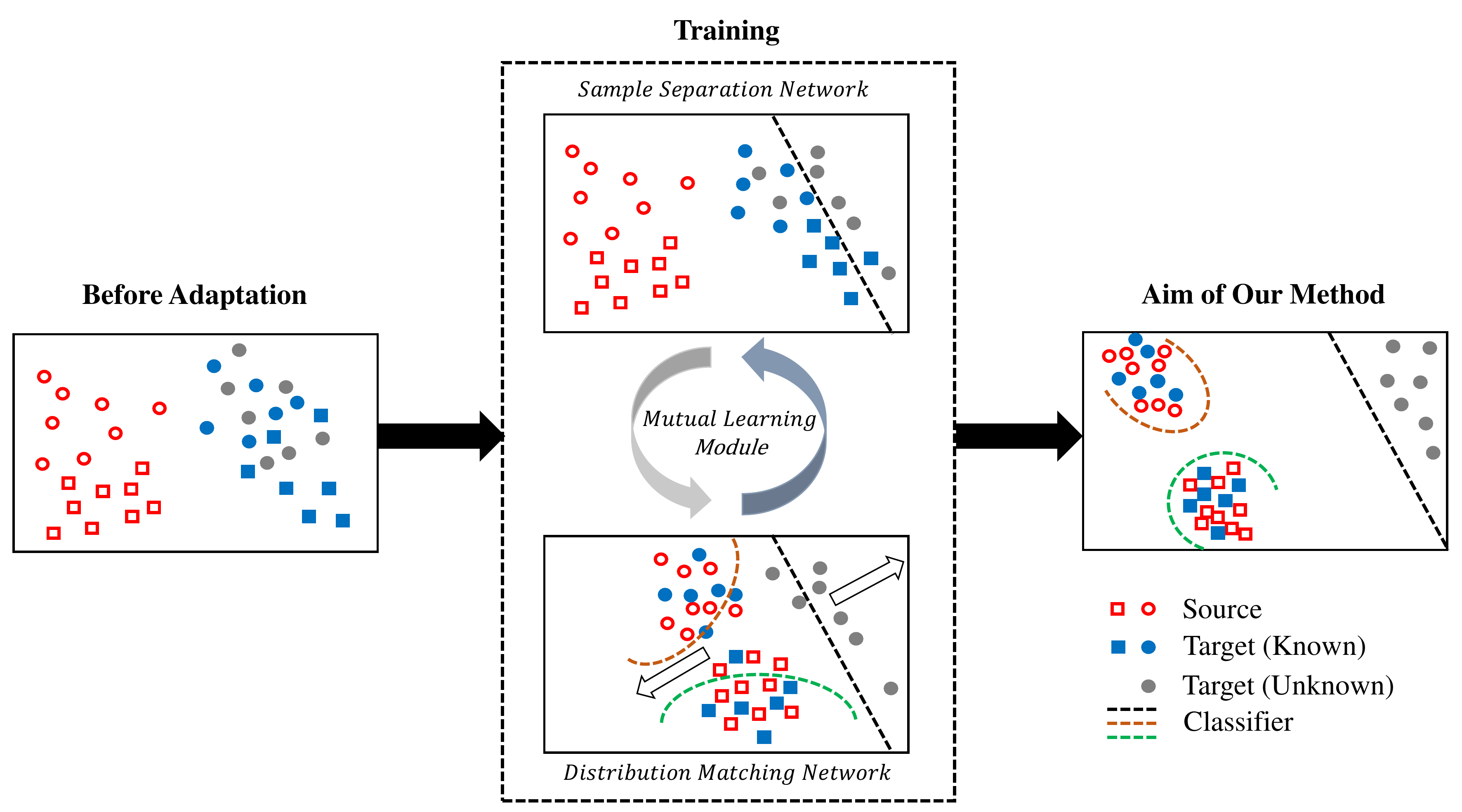}
\end{center}
   \caption{An overview of proposed Mutual to Separate approach to open set domain adaptation. {Red color indicates source domain, blue and grey colors represent target domain. Different shapes point to different classes.}
  {During training, the~\emph{Sample Separation Network (SSN)} is trained to learn a hyperplane for separating unknown samples from known ones, while the~\emph{Distribution Matching Network (DMN)} is trained to  match domain distributions and push apart unknown samples. Crucially, those two networks benefit each other via the~\emph{Mutual Learning Module}.~\emph{Best viewed in color}.}}
\label{fig:overview}

\end{figure*}

In an attempt to address the domain gap, the domain adaption (DA) literature has continuously been relaxing the target domain labeling constraints. It started with a supervised setting~\cite{daume2009frustratingly,bergamo2010exploiting,aytar2011tabula} where training relies on the target domain data being fully labeled, followed by semi-supervised~\cite{tzeng2015simultaneous,yao2015semi} where only partial labeling is present, and then converged at unsupervised~\cite{xu2018deep,chen2019progressive,saito2018maximum} that completely removes all labeling constraints. Early unsupervised domain adaption methods mostly assume a ``closest set'' setting, where the source and target domains share the same label space. These methods generally seek to bridge the domain gap by \textit{distribution matching} at feature-level~\cite{ganin2014unsupervised,liu2019separate,Kaichao2019Universal,long2015learning,tzeng2017adversarial} or pixel-level~\cite{sankaranarayanan2018generate,hoffman2018cycada,hu2018duplex,murez2018image,liu2018detach}. A newly emerging stream of research has moved onto an even harder yet more practical ``open set'' setting, where there is no assumption of any label information from target domain data, nor any knowledge on the unknown classes.
Very recent studies~\cite{liu2019separate,saito2018open,panareda2017open,bendale2016towards,tan2019weakly,feng2019attract,Kaichao2019Universal} have shown to yield performances exceeding that of closed set domain adaption. 


This paper adopts the open set domain adaption (OSDA) setting, yet with a new initiative in pushing the conventional domain gap boundary that prior research had become accustomed to -- we seek an OSDA solution that would generalize well under more drastic domain shifts. We start by highlighting different degrees of domain shifts under commonly used datasets (Office-$31$~\cite{saenko2010adapting}, Office-Home~\cite{venkateswara2017deep}), and the one that we for the first time re-purposed for OSDA (PACS~\cite{Li2017dg}). This can be observed in Figure~\ref{fig:OSDA:1a}, where it shows increasing domain shifts  from Office-$31$~\cite{saenko2010adapting} (photos from just offices) to Office-Home~\cite{venkateswara2017deep} (photos from both office and home) with PACS \cite{Li2017dg} exhibiting the largest domain shift (photos and sketches).
We further observe that in the presence of larger domain gaps, existing OSDA solutions would witness a significant performance drop. Moreover, the degree of this drop is directly proportional to that of performance shift -- the larger the shift, the higher the performance degradation (Figure~\ref{fig:OSDA:1b}). This is also intuitive from the feature distribution plots in Figure~\ref{fig:OSDA:1a}: for Office-$31$, known and unknown features have a clean separation, and source and target distributions are well aligned; PACS being at the other end of the extreme, exhibits much more convoluted known and unknown data, and very weakly correlated source and target distributions.


Two key factors need to be considered while addressing domain gap in OSDA: (i) correctly classifying data of all unknown classes as ``unknown'', and (ii) matching domain distributions between source and target domains in the shared label space. Most existing methods employ only one model to simultaneously achieve both~\cite{saito2018open,panareda2017open,Kaichao2019Universal,feng2019attract}. It was not until very recently,~\cite{liu2019separate} demonstrated that it is beneficial to execute (i) and (ii) separately under two disjoint networks, which had led to state-of-the-art performances. The \textit{key intuition} behind this paper however, is that under such a single model, (i) and (ii)  would potentially inflict negative transfer on each other. 
This is because, under large domain shifts, known and unknown samples become more confused in the target domain, making it harder to learn an accurate hyperplane to separate them. This would lead to wrongly classified unknown samples, adversely influencing subsequent domain-distribution matching. Consequently, after domain matching, unknown samples will be confused further with known data, making the hyperplane even more difficult to learn.

In this paper, instead of avoiding negative transfer as such~\cite{liu2019separate}, we focus on the opposite -- how to cultivate positive mutual exchange between these two tasks, with the hope that such information exchange can better accommodate for larger domain shifts. The key challenge is therefore how to encourage positive transfer between the two key tasks. To this end, we propose \textit{Mutual to Separate} (MTS), a deep mutual learning approach for OSDA. Figure~\ref{fig:overview} offers an overview, where MTS trains two networks mutually: (i) the Sample Separation Network (SSN) that learns a hyperplane for separating unknown samples in the target domain from known samples, and (ii) the Distribution Matching Network (DMN) that enables positive information exchange by distancing \textit{unknown} data samples in addition to matching domain distributions between source and target domains.
It follows that these two networks when coupled via a mutual learning module can positively benefit each other: with the help of better classified ``known'' samples obtained from SSN, DMN can better estimate the underlying domain distributions; and upon finishing its iteration, DMN would inform SSN on the refined state of unknown samples.
By cultivating this mutually beneficial information exchange, MTS is able to tackle larger domain gaps. As seen in Figure~\ref{fig:OSDA:1b}, our method outperforms other state-of-the arts, most significantly on PACS \cite{Li2017dg} where the domain gap is largest, along with Office-$31$~\cite{saenko2010adapting} and Office-Home~\cite{venkateswara2017deep} datasets.

The key contributions of this paper can be summarized as: (i) a novel mutual learning setup specifically designed for OSDA under large domain shifts, (ii) a dual network configuration specifically designed to enable positive information exchange, and (iii) PACS~\cite{Li2017dg} re-purposed for OSDA as a dataset that exhibits larger domain shifts. Extensive experiments are carried out on two commonly used OSDA datasets (Office-$31$~\cite{saenko2010adapting}, Office-Home~\cite{venkateswara2017deep}) along with the newly re-purposed PACS~\cite{Li2017dg}. Results show that our model can outperform the current state-of-the-arts by a significant margin. Ablative studies are further conducted to draw insights towards each of the design choices.


\section{Related Work}

\subsection{Mutual Learning} 
With the aim of acquiring training experience from another network, distillation based methods~\cite{hinton2015distilling} were proposed to train a relatively small separate network.
{However, mutual learning has proven to be more efficient for cultivating information exchange between networks.}
Unlike distillation, mutual learning starts with a collection of essential networks, learning jointly to complete its objectives. Batra \emph{et al.} introduced a similar cooperative learning approach~\cite{batra2017cooperative}, where various models specializing in various domains were jointly trained to understand domain-invariant visual attributes. Zhang \emph{et al.} on the other hand, put forth a deep mutual learning model~\cite{zhang2018deep} that reduces the divergence between outputs from two networks having different parameter initialization and dropouts. With the aim of enhancing mutual learning using an advanced teacher model, Tarvainen \emph{et al.}~\cite{tarvainen2017mean} asserted exponential moving-average of a student network, as a teacher to cite training targets for the student.

On the contrary, our mutual learning module is based on two networks mutually enhancing each other using information transferred between them.
To the best of our knowledge, capturing such interaction information between separate networks to enhance mutual learning, has not been attempted earlier.

\subsection{Open Set Domain Adaptation} 
Quite a few studies have surfaced related to this emerging hot-spot of a topic in the computer vision community. For example, distance from every target sample's feature to every source class's center is used in Assign-and-Transform-Iteratively~\cite{panareda2017open} (ATI) to determine the class of the target sample.
A feature generator is trained in an adversarial training framework, in Open Set Back-Propagation (OSBP)~\cite{saito2018open} to deviate the probabilistic value of a target sample to be classified as “unknown”, from its pre-defined threshold.
Later, Long~\emph{et al.} separates samples of unknown classes from known ones, matching features of known-class samples across source and target domains via a progressive mechanism~\cite{liu2019separate}. Similarly, both prediction uncertainty and domain similarity of every sample is utilized in Universal Domain Adaptation (UAN)~\cite{Kaichao2019Universal} to develop a weighting mechanism for discovering label sets shared by both domains, thus promoting common-class adaptation. Other works include Qianyu \emph{et al.}~\cite{feng2019attract}, using semantic categorical alignment to achieve proper separability of target known classes and semantic contrasting mapping, to distance the unknown class from the decision boundary. However, for all such methods, problems arise when domain shifts significantly, especially varying largely on the lower side.

To conclude, we develop a  Mutual to Separate (MTS) approach to address the concept of open set domain adaptation. Akin to earlier methods~\cite{liu2019separate,zhang2018deep,ganin2014unsupervised}, our MTS employs a multi-binary classifier with a domain adversarial network to separate unknown samples, while adapting source and target domains in the shared label space simultaneously. Furthermore, these two challenges mutually improve each other via a novel mutual learning module, additionally proving its robustness to a variety of domain shifts.

\section{Method}

\subsection{Open Set Domain Adaptation}
Inspired from~\cite{liu2019separate,Kaichao2019Universal,saito2018open}, we define the source domain as 
$D_s = \{x_i^s,y_i^s\}_{i=1}^{n_s}$ having $n_s$ labeled examples, and a target domain $D_t = \{x_j^t\}_{j=1}^{n_t}$ of $n_t$ unlabeled examples. The source domain contains a set of classes $C_s$, which is common to the target domain, $C_t$, \textit{i.e.,} $C_s \subset C_t$. In addition to $C_s$, $C_t$ also consists of \textit{unknown} classes referred to as $C_{u}$. Essentially, $C_t = C_s \cup C_u$.

\begin{figure*}[t]
\begin{center}
  \includegraphics[width=1.0\linewidth]{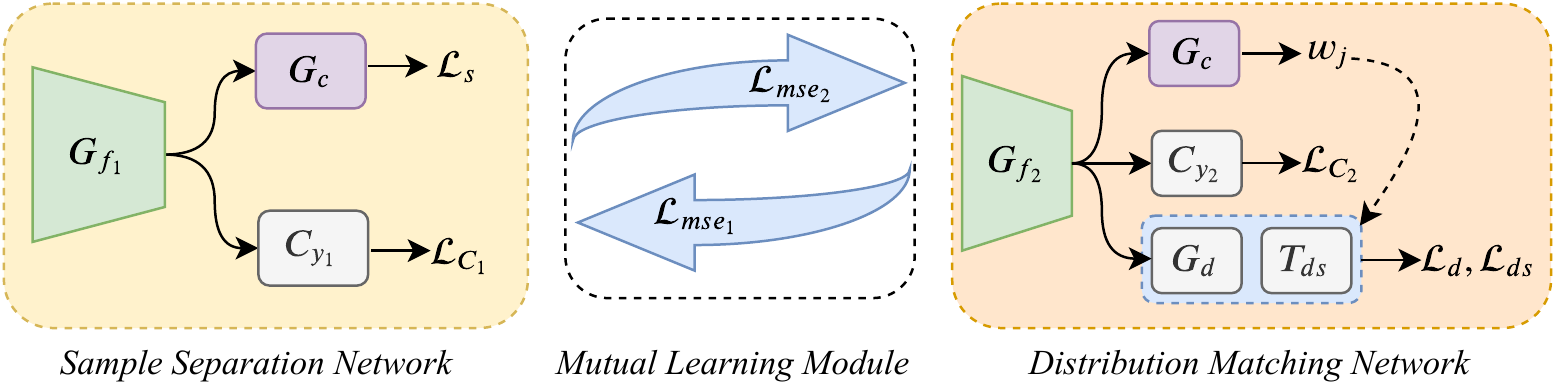}
\end{center}
  \caption{The proposed \textbf{Mutual to Separate} approach for open set domain adaptation, which contains three components:~\emph{a Sample Separation Network (SSN)}, 
  ~\emph{a Distribution Matching Network (DMN)}, and ~\emph{a Mutual Learning Module}. 
  SSN consists of a feature extractor $G_{f_1}$, a classifier $C_{y_1}$, and  a multi-binary classifier $G_{c}$ to learn hyperplanes for each class.
  DMN consists of: a feature extractor $G_{f_2}$, an extended classifier $C_{y_2}$, a multi-binary classifier $G_{c}$ to generate the weights $w_j$ for rejecting target samples belonging to the unknown classes, a domain discriminator $G_d$ to perform adversarial domain adaptation between source and target domains in the shared label space, and a domain separating classifier $T_{ds}$ to distance the unknown samples in target domain during domain adaptation. Parameters are shared by $G_{c}$ in different networks.} 

\label{fig:architecture}

\end{figure*}



\subsection{Mutual to Separate}

The main challenges of open set domain adaptation include separation of unknown samples in target domain and reduction of domain shift between source and target domains in the shared label space. 
It has been observed that these challenges can influence each other. 
Therefore, a logical approach towards solving this is understanding how to cultivate positive mutual exchange between these two tasks.
Using this idea we design our architecture, as shown in Figure~\ref{fig:architecture}.
Details of components used in our proposed method are described below.

\subsubsection{Sample Separation Network (SSN)}

This network aims to learn a hyperplane for each class in $C_s$ thus separating unknown samples from known ones. Being similar to a state-of-the-art approach \cite{liu2019separate} in this regard, it consists of a feature extractor $G_{f_1}$, a classical classifier $C_{y_1}$ for $|C_s|$ classes, and a multi-binary classifier $G_c$.  
Here $C_{y_1}$ is trained with the sole motivation of recognizing source domain samples, thus we define the classification loss ($\mathcal{L}_{C_1}$) as:
\begin{equation}
{\mathcal{L}_{C_1}} = {\frac{1}{n_s} {{\sum_{i=1}^{n_s}}{{\mathcal{L}_{CE}}(C_{y_1}(G_{f_1}(x_i^s)), y_i^s)}}},
\end{equation}
where ${\mathcal{L}_{CE}}$ is cross-entropy loss.

Thereafter, this network is run with a focus on separating known and unknown samples, followed by classifying unknown classes into a separate category from known ones, in the target domain.
For the latter part, we employ a multi-binary classifier as the weighting scheme~\cite{liu2019separate}, to measure the similarity between each target sample and corresponding source class. The loss for all classifiers can be defined as:

\begin{equation}
\mathcal{L}_s = {\frac{1}{|{C_s}|}} {\sum_{c=1}^{|{C_s}|}{\frac{1}{n_s}} \sum_{i=1}^{n_s}{{\mathcal{L}_{BCE}}(G_{c}({G_{f_1}}(x_i^s)), I(y_i^s, c)}})   ,
\\
\\
\label{equation:ls}
\end{equation}
where $\mathcal{L}_{BCE}$ is the binary cross-entropy loss, and 
${I(y_i^s, c)} = 1_{{y_i^s = c}}$.
The overall loss function $\mathcal{L}_{\mathbf{\Theta}_1}$ for SSN network $\mathbf{\Theta}_1 (\theta_{f_1},\theta_{y_1},
\theta_{c}{|_{c=1}^{|C_s|}}  )$ is defined as:

\begin{equation}
\mathcal{L}_{\mathbf{\Theta}_1} = {\mathcal{L}_{C_1}} + {\mathcal{L}_s},
\end{equation}
where $\theta_{f_1},\theta_{y_1}$, and $\theta_{c}{|_{c=1}^{|C_s|}} $ denote the parameters of $G_{f_1}$, $C_{y_1}$, and $G_{c}{|_{c=1}^{|C_s|}}$ respectively.

\subsubsection{Distribution Matching Network (DMN)}
This network consists of a feature extractor $G_{f_2}$, an extended classifier $C_{y_2}$, a multi-binary classifier $G_c$, a domain discriminator $G_d$ to perform adversarial domain adaptation between source and target domains in the shared label space~\cite{liu2019separate,feng2019attract,saito2018open}, and a domain separation classifier $T_{ds}$ to drive apart the unknown samples in the target domain during domain adaptation.

Specifically, for every target sample, each binary classifier $G_{c} (c \in [1,|C_s|])$, outputs $p_c$ as the probability of that sample belonging to the known class $c$. Hence $p_c$ can be explained as the similarity between the target sample and known class $c$. We use the highest probability in $\{p_1, p_2,~\cdots, p_{|{C_s}|}\}$ as the measure of similarity ($w_j$) between each target sample $x_j^t$ and the source domain:

\begin{equation}
w_j =  {\max_{c \in [1,|C_s|]}} {{G_{c}}({G_{f_2}}({x_j^t}))} ,
\label{equation:wj2}
\end{equation}
where $G_{f_2}$ is a feature extractor in the DMN.
Therefore, the classification loss ($\mathcal{L}_{C_2}$) between the source domain samples and the most ``unknown'' sample in target domain for this network can be defined as follows,

\begin{equation}
\begin{aligned}
\mathcal{L}_{C_2}&\qquad  = 
{{\frac{1}{n_s} {{\sum_{i=1}^{n_s}}{{\mathcal{L}_{CE}}({C_{y_2}^{1:{|C_s|}}(G_{f_2}(x_i^s))}, y_i^s)}}}}
\\ &\qquad + {{ {w_j}{\mathcal{L}_{CE}(C_{y_2}^{{|C_s|+1}}(G_{f_2}(x_j^t)),l_{uk})}}},
\end{aligned}
\end{equation}
%
where~$\textit{l}_{uk}$ is the label of unknown class, $C_{y_2}$ is an extended classifier for $|C_s| + 1$ classes, having $|C_s|$ classes from the source domain and $1$ extra as the class of ``unknown" samples.
We especially chose only the most ``unknown'' samples in target domain to train the classification loss based on ${w_j}$. 

Thereafter, we have focused our model on aligning the distributions of source and target data in the shared label space $C_s$. Without using hard discrimination, we exploit similarity ${w_j}$ between each target sample and the source domain as a soft instance-level weight, where higher $w_j$ indicates a higher chance of being from a known class. This helps us define a \textit{weighted} adversarial adaptation loss ($\mathcal{L}_{d}$) for feature distributions in the shared label space $C_s$ as:



\begin{equation}
\begin{aligned}
\mathcal{L}_d &\qquad= 
{{\frac{1}{n_s}{\sum_{i=1}^{n_s}}{\mathcal{L}_{BCE}({G_d(G_{f_2}(x_i^s))},1)}}}
\\ &\qquad + {{\frac{1}{\sum_{j=1}^{n_t}{w_j}}}{\sum_{j=1}^{n_t}}{w_j} {\mathcal{L}_{BCE}(1 - {G_d(G_{f_2}(x_j^t))},1)}},
\end{aligned}
\end{equation}
%
%
%
%
%
where $G_d$ is a domain discriminator performing adversarial domain adaptation between source and target domains in the shared label space.



To distance the unknown {samples} further from samples in source domain as well as from known samples in target domain, we utilize a multi-binary classifier, which is composed of three binary classifiers, denoted by $T_{ds}|^3_{ds=1}$.
{
$Class_1$ contains samples belonging to $D_s$; 
$Class_2$ holds samples belonging to the \textit{known} classes in $D_t$ while 
$Class_3$ has samples belonging to the \textit{unknown} classes in $D_t$.
}

Randomly selecting one sample from the $D_s$ classifies it to be in $Class_1$, represented by ($\pi_1,1$). 
With reference to Equation~\ref{equation:wj2}, $w_j$ can indicate the similarity between each target sample $x_j^t$ and the source domain ($D_s$). 
Thus, we rank the similarity for the target samples in any mini-batch, and classify the sample with highest similarity to be in $Class_2$, represented by ($\pi_2,2$).
Similarly, the sample with lowest similarity is classified to be in $Class_3$, represented by ($\pi_3,3$), thus providing three samples overall, $\{\pi_i, i\}_{i=1}^{3}$.
The loss for three classifiers can be defined as:
\begin{equation}
\mathcal{L}_{ds} = {\frac{1}{3}}{\sum_{ds=1}^{3}{\frac{1}{3}} \sum_{i=1}^{3}{{\mathcal{L}_{BCE}}({T_{ds}}{({G_{f_2}}(\pi_i))}, {I(i, ds)}})},
\end{equation}
where ${I(i, ds)} =1_{i = ds}$.



In this section, we implement adversarial adaptation to {match domain distribution} between $D_s$ and $D_t$ in the shared label space. Therefore, the overall loss function consists of two parts namely $\mathcal{L}_{\mathbf{\Theta}_{2a}}$ and $\mathcal{L}_{\mathbf{\Theta}_{2b}}$. $\mathcal{L}_{\mathbf{\Theta}_{2a}}$ for DMN network $\mathbf{\Theta}_{2} (\theta_{y_2}, \theta_d,  \theta_{ds}{|}{_{ds=1}^3} )$ is defined as:

\begin{equation}
{\mathcal{L}_{\mathbf{\Theta}_{2a}}} ={ {\mathcal{L}_{C_2}} +  {\mathcal{L}_d}  + \alpha {\mathcal{L}_{ds}}  },
\end{equation}
where $\theta_{y_2},\theta_d$, and $\theta_{ds}{|}{_{ds=1}^3}$ denote the parameters of $C_{y_2}$, $G_{d}$, and $T_{ds}{|}{_{ds=1}^3}$ respectively, and $\alpha$ is the hyper-parameter to trade off the entropy loss.
Similarly, $\mathcal{L}_{\mathbf{\Theta}_{2b}}$ for DMN network $\mathbf{\Theta}_{2} (\theta_{f_2}, \theta_{ds}{|}{_{ds=1}^3} )$ is defined as
\begin{equation}
{\mathcal{L}_{\mathbf{\Theta}_{2b}}} ={ {\mathcal{L}_{C_2}}  - {\mathcal{L}_d}  + \alpha{\mathcal{L}_{ds}} },
\label{equation:domain_2}
\end{equation}
where $\theta_{f_2} $ denotes the parameters of $G_{f_2}$; $\alpha$ is the hyper-parameter to trade off the entropy loss.  It is worth mentioning that in this scenario, we may generate domain-invariant features by using Equation~\ref{equation:domain_2}.
Therefore, the ${I(i, ds)}$ in ${\mathcal{L}_{ds}}$ should revise the true label of $Class_1$ and $Class_2$
as ${I(i, ds)} = {1}_{i,ds \neq 3}  (i,ds \in {1,2,3})$ except that $I(3,3)=1$.
We no longer distinguish between $class_1$ and $class_2$, but enhance the differences between $class_1$ ($class_2$) and $class_3$ only, thus being more favorable to SSN for identification of unknown samples, and beneficial for DMN to generate domain-invariant features.

\subsubsection{Enhanced Mutual Learning Module}

We had proposed that the mutual learning module is built on two different specializing tasks networks. It captures the interaction information between separate networks by minimizing the divergence between the two essential networks, thus improving them both.

As mentioned earlier, SSN can separate known/unknown samples in target domain, and DMN can match domain distribution between source and the \textit{known} part of target domain in the shared label space, while separating features of unknown classes farther apart. Therefore, we use the information provided by both networks, to boost ${G_{c}}$'s ability of separating unknown samples, and ${G_{d}}$'s ability of matching domain distribution, between source domain and target domains in the shared label space.
For the $\mathrm{SSN}$ this can be represented as:
%
%
\begin{equation}
{\mathcal{L}_{mse_1}}=
\frac{1}{2} \bigl(\frac{1}{n_s} \sum_{i=1}^{n_s}{(c^{s_1}_i - c^{s_2}_i)^2} + \frac{1}{n_t} \sum_{j=1}^{n_t}{(c^{t_1}_j - c^{t_2}_j)^2}\bigr) ,
\end{equation}
where ${c^{s_1}_i} = {G_{c}}{({G_{f_1}}(x_i^s))}$, ${c^{s_2}_i} = {G_{c}}{({G_{f_2}}(x_i^s))}$,
${c^{t_1}_j} = {G_{c}}{({G_{f_1}}(x_j^t))}$, and ${c^{t_2}_j} = {G_{c}}{({G_{f_2}}(x_j^t))}$. 

Similarly for DMN, we have:
\begin{equation}
\mathcal{L}_{mse_2} = \frac{1}{2} \bigl(\frac{1}{n_s} \sum_{i=1}^{n_s}{(c^{s_2}_i - c^{s_1}_i)^2} + \frac{1}{n_t} \sum_{j=1}^{n_t}{(c^{t_2}_j-c^{t_1}_j)^2}\bigr).
\end{equation}


\begin{table*}[!t]
    \caption{Classification Accuracy (\%) of OSDA tasks on Office-$31$ (ResNet-$50$). 
    }
       \begin{adjustbox}{width=1.0\linewidth,center}
    \begin{tabular}{ccccccccccccccc}
    \hline
    \multirow{2}*{Method} & \multicolumn{2}{c}{A--W} & \multicolumn{2}{c}{A--D} & \multicolumn{2}{c}{D--W} & \multicolumn{2}{c}{W--D} & \multicolumn{2}{c}{D--A} & \multicolumn{2}{c}{W--A} & \multicolumn{2}{c}{Avg.} \\
\cline{2-15}          & OS    & OS*   & OS    & OS*   & OS    & OS*   & OS    & OS*   & OS    & OS*   & OS    & OS*   & OS    & OS* \\
    \hline
    \hline
    ResNet50 (CVPR16)~\cite{he2016deep}  & 82.5$\pm$1.2 & 82.7$\pm$0.9 & 85.2$\pm$0.3 & 85.5$\pm$0.9 & 94.1$\pm$0.3 & 94.3$\pm$0.7 & 96.6$\pm$0.2 & 97.0$\pm$0.4 & 71.6$\pm$1.0 & 71.5$\pm$1.1 & 75.5$\pm$1.0 & 75.2$\pm$1.6 & 84.2  & 84.4 \\
    RTN (NeurIPS16)~\cite{long2016unsupervised}    & 85.6$\pm$1.2 & 88.1$\pm$1.0 & 89.5$\pm$1.4 & 90.1$\pm$1.6 & 94.8$\pm$0.3 & 96.2$\pm$0.7 & 97.1$\pm$0.2 & 98.7$\pm$0.9 & 72.3$\pm$0.9 & 72.8$\pm$1.5 & 73.5$\pm$0.6 & 73.9$\pm$1.4 & 85.4  & 86.8 \\
    DANN (ICML15)~\cite{ganin2014unsupervised}  & 85.3$\pm$0.7 & 87.7$\pm$1.1 & 86.5$\pm$0.6 & 87.7$\pm$0.6 & \tblue{97.5$\pm$0.2} & 98.3$\pm$0.5 & \tred{99.5$\pm$0.1} & \tred{100.0$\pm$.0} & 75.7$\pm$1.6 & 76.2$\pm$0.9 & 74.9$\pm$1.2 & 75.6$\pm$0.8 & 86.6  & 87.6 \\
    OpenMax (CVPR16)~\cite{bendale2016towards} & 87.4$\pm$0.5 & 87.5$\pm$0.3 & 87.1$\pm$0.9 & 88.4$\pm$0.9 & 96.1$\pm$0.4 & 96.2$\pm$0.3 & 98.4$\pm$0.3 & 98.5$\pm$0.3 & 83.4$\pm$1.0 & 82.1$\pm$0.6 & 82.8$\pm$0.9 & 82.8$\pm$0.6 & 89.0    & 89.3 \\
    ATI-$\lambda$ (ICCV17)~\cite{panareda2017open} & 87.4$\pm$1.5 & 88.9$\pm$1.4 & 84.3$\pm$1.2 & 86.6$\pm$1.1 & 93.6$\pm$1.0 & 95.3$\pm$1.0 & 96.5$\pm$0.9 & 98.7$\pm$0.8 & 78.0$\pm$1.8 & 79.6$\pm$1.5 & 80.4$\pm$1.4 & 81.4$\pm$1.2 & 86.7  & 88.4 \\
    OSBP (ECCV18)~\cite{saito2018open}   & 86.5$\pm$2.0 & 87.6$\pm$2.1 & 88.6$\pm$1.4 & 89.2$\pm$1.3 & 97.0$\pm$1.0 & 96.5$\pm$0.4 & 97.9$\pm$0.9 & 98.7$\pm$0.6 & 88.9$\pm$2.5 & 90.6$\pm$2.3 & 85.8$\pm$2.5 & 84.9$\pm$1.3 & 90.8  & 91.3 \\
    STA (CVPR19)~\cite{liu2019separate} &\tblue{89.5$\pm$0.6} & 92.1$\pm$0.5 & \tblue{93.7$\pm$1.5} & 96.1$\pm$0.4 & \tblue{97.5$\pm$0.2} & 96.5$\pm$0.5 & \tblue{99.5$\pm$0.2} & \tblue{99.6$\pm$0.1} & \tblue{89.1$\pm$0.5} & \tblue{93.5$\pm$0.8} & \tblue{87.9$\pm$0.9} & 87.4$\pm$0.6 & \tblue{92.9}  & 94.1\\
    UAN (CVPR19)~\cite{Kaichao2019Universal}  & 85.6$\pm$1.1 & \tblue{93.6$\pm$1.4} & 89.9$\pm$0.8 & \tred{98.5$\pm$0.8} & 92.6$\pm$0.0 & \tred{99.8$\pm$0.2} & 92.0$\pm$0.4 & \tred{100.0$\pm$0} & 87.9$\pm$0.1 & \tred{95.2$\pm$0.2} & 87.5$\pm$0.1 & \tred{95.8$\pm$0.0} & 89.3  & \tred{97.1} \\
    \hline
    \hline
    Ours  & \tred{92.4$\pm$0.3} & \tred{96.8$\pm$0.8} & \tred{94.7$\pm$0.2} & \tblue{98.2$\pm$0.5} & \tred{97.9$\pm$0.2} & \tblue{99.5$\pm$0.2} & 98.9$\pm$0.4 & \tred{100.0$\pm$0} & \tred{89.6$\pm$0.4} & 92.0$\pm$0.3 & \tred{89.7$\pm$0.2} & \tblue{91.9$\pm$0.3} & \tred{93.8} & \tblue{96.4} \\
    \hline
    \end{tabular}%
\end{adjustbox}

  \label{tab:office31}%
\end{table*}%
  
\subsubsection{Optimization}

Throughout the entire training process, mutual learning strategy is implemented at every updation step of the mini-batch based model. For every iteration, we compute loss function of the two networks and update their parameters. This optimization of $\mathbf{\Theta}_1$ and $\mathbf{\Theta}_2$ is conducted iteratively until convergence. The optimization details are summarized as follows:

\textbf{Step 1.} We first train SSN to classify source samples. Meanwhile, the multi-binary classifier $G_{c}{|_{c=1}^{|C_s|}}$ is trained in a one-vs-rest way for each source class.
The optimal parameters $\hat{\theta}_{f_1}, \hat{\theta}_{y_1}$, and $\hat{\theta}_{c}{|_{c=1}^{|C_s|}}$ can be found by:

\begin{equation}
   ({\hat{\theta}_{f_1}, \hat{\theta}_{y_1}, \hat{{\theta}}_{c}{|_{c=1}^{|C_s|}}})= 
   {\mathop{\arg\min}_{\theta_{f_1}, \theta_{y_1}, \theta_{c}{|_{c=1}^{|C_s|}}} {\mathcal{L}_{\mathbf{\Theta}_1} + \beta {\mathcal{L}_{mse_1}} }},
\end{equation}
where $\beta$ is a hyper-parameter to trade off $\mathcal{L}_{\mathbf{\Theta}_1}$.

\textbf{Step 2.} 
Here, we train DMN to generate domain-invariant feature representation for source and target domain in the shared label space, and train $T_{ds}{|}{_{ds=1}^3}$ to distance the unknown sample further from both the source domain sample and target known sample.
Having $\beta$ as a hyper-parameter to trade off $\mathcal{L}_{\mathbf{\Theta}_{2a}}$ and $\mathcal{L}_{\mathbf{\Theta}_{2b}}$, the optimal parameters $\hat{\theta}_{f_2}, \hat{\theta}_{y_2}, \hat{\theta}_d$, and $\hat{\theta}_{ds}{|}{_{ds=1}^3} $ can be found by:

\begin{equation}
({\hat{\theta}_{y_2}, \hat{\theta}_{d}, \hat{\theta}_{ds}{|}{_{ds=1}^3}})= 
{\mathop{\arg\min}_{\theta_{y_2}, \theta_{d}, \theta_{ds}{|}{_{ds=1}^3}} {\mathcal{L}_{\mathbf{\Theta}_{2a}} + \beta {\mathcal{L}_{mse_2}} }} 
\end{equation}

\begin{equation}
({\hat{\theta}_{f_2}, \hat{\theta}_{ds}{|}{_{ds=1}^3}})= 
{\mathop{\arg\min}_{\theta_{f_2}, \theta_{ds}{|}{_{ds=1}^3}} {\mathcal{L}_{\mathbf{\Theta}_{2b}} + \beta {\mathcal{L}_{mse_2}} }}.
\end{equation}


\section{Experiments}\label{ALLExperiments}
\subsection{Datasets}
In addition to two commonly used OSDA benchmark datasets, we introduce a re-purposed OSDA benchmark with even larger domain shifts, for a comprehensive evaluation of our method.
The datasets are: (i)~\textbf{Office-$31$}~\cite{saenko2010adapting}, which is a standard benchmark for domain adaptation in computer vision with three domains Amazon (\textbf{A}), Webcam (\textbf{W}) and DSLR (\textbf{D}); (ii)~\textbf{Office-Home}~\cite{venkateswara2017deep} which is an equally challenging domain adaptation dataset, and is collected in a similar manner.
(iii)~\textbf{PACS}: For the final dataset, we re-purpose the PACS~\cite{Li2017dg} benchmark which is another such difficult domain adaptation dataset, to evaluate OSDA.
PACS holds 4 different domains namely: photo (\textbf{Ph}), sketch (\textbf{Sk}), cartoon (\textbf{Ca}), and art\_painting (\textbf{Ar}). This has both higher practical relevance, and exhibits higher domain shift than existing benchmarks. Each domain contains images from $7$ object classes. 

\subsection{Competitors} 
We compare our proposed method with several open set recognition, domain adaptation, and open set domain adaptation methods~\cite{liu2019separate} for a comparative evaluation as discussed ahead.
Open Set SVM (\textbf{OSVM})~\cite{jain2014multi} is an SVM based method that uses thresholding for each class to recognize samples and discard outliers.
\textbf{MMD + OSVM} and \textbf{DANN + OSVM} are two variants of OSVM instilling Maximum Mean Discrepancy~\cite{gretton2007kernel} and domain adversarial network~\cite{ganin2014unsupervised} in OSVM respectively. \textbf{ATI-$\lambda$+OSVM} maps the feature space of source domain to the target domain by assigning images in the latter to known categories~\cite{panareda2017open}. While \textbf{OSBP} uses an adversarial classifier to deal with samples of unknown classes~\cite{saito2018open}, \textbf{OpenMAX} is a deep open set recognition approach which has a module structured for discarding outliers~\cite{bendale2016towards}. 
One of the most recent domain adaptation method is \textbf{STA} that achieved state-of-the-art performance using a set of binary classifiers along with an adversarial classifier to handle samples of unknown classes~\cite{liu2019separate}. \textbf{UAN} on the other hand is a universal method, that aims to solve the problem of universal domain adaptation~\cite{Kaichao2019Universal}.
Other relevant methods include \textbf{DANN}~\cite{ganin2014unsupervised} and \textbf{RTN}~\cite{long2016unsupervised}.
For closed-set methods, we use a confidence threshold to judge if a sample is from the unknown classes or not. We follow common evaluation protocols for each dataset. \textit{Note: for all tables, the best and second best scores are marked in red and blue respectively.}

\subsection{Evaluation on Office-$31$ dataset}\label{subsection:office31}

\subsubsection{Setting}  
Office-$31$ contains $4,652$ images from $31$ categories.  
Inspired from earlier works~\cite{saito2018open,liu2019separate}, we select $10$ classes as shared classes, which are in common with the Caltech dataset~\cite{gong2012geodesic}. Ordered in an alphabetical order, $21$-$31$ classes are used for unknown samples in the target domain, whereas classes $11$-$20$ are usually used for unknown samples in the source domain, which being unnecessary was not used for our method. These tasks denote the performance where the source and target domains have small domain gap. For a fair comparison, we follow~\cite{liu2019separate} to use ResNet-$50$ as a backbone along with a domain adversarial network similar to DANN~\cite{ganin2014unsupervised}.  Furthermore, we employ two evaluation metrics: 
~\textbf{OS}: normalized accuracy for all classes including the unknown as one class, OS $ =\frac{1}{K + 1} \sum_{k=1}^{K+1}{\alpha_k} $, where $K$ indicates number of known classes, $\alpha_k$ signifies accuracy of the $\mathrm{k^{th}}$ class and $K$+$1^{th}$ class denotes the unknown class.; ~\textbf{OS*}: normalized accuracy only on known classes. OS* $=\frac{1}{K } \sum_{k=1}^{K}{\alpha_k} $; 
We use a momentum-SGD optimizer with a learning rate of $10^{-4}$, momentum $0.9$, weight decay $5\times10^{-4}$ and set $\alpha$=$0.8$ and $\beta$=$0.5$. For our experiment, we execute each method thrice, noting the average accuracy and standard deviation values. 

\subsubsection{Results}
From results in Table~\ref{tab:office31}, it can be concluded that: 
(i) Closed set domain adaptation methods achieve lower performance than ResNet on a few tasks. Such methods work unsatisfactorily even with confidence thresholding. This sink in performance is caused by negative transfer due to incorrect matching of unknown classes in target domain with the known ones in source domain.
(ii) Using different evaluation metrics, our method has evidently outperformed the others on most tasks. 
(iii) Speaking comprehensively, our method achieves the best performance, improving by $9.6$\% (OS) on ResNet-$50$, and by $0.9$\% (OS) on prior state-of-the-art STA.

\begin{table*}[!t]

    \caption{Classification accuracy OS(\%) of OSDA tasks on Office-Home (ResNet-$50$).
    }
      \begin{adjustbox}{width=1.0\linewidth,center}
    \begin{tabular}{cccccccccccccc}
    \hline
    Method & Ar-Cl & Pr-Cl & Rw-Cl & Ar-Pr & Cl-Pr & Rw-Pr & Cl-Ar & Pr-Ar & Rw-Ar & Ar-Rw & Cl-Rw & Pr-Rw & Avg. \\
    \hline
    \hline
    ResNet50 (CVPR16)~\cite{he2016deep} & 53.4$\pm$0.4 & 52.7$\pm$0.6 & 51.9$\pm$0.5 & 69.3$\pm$0.7 & 61.8$\pm$0.5 & 74.1$\pm$0.4 & 61.4$\pm$0.6 & 64.0$\pm$0.3 & 70.0$\pm$0.3 & 78.7$\pm$0.6 & 71.0$\pm$0.6 & 74.9$\pm$0.9 & 65.3 \\
    ATI-$\lambda$ (ICCV17)~\cite{panareda2017open}    & 55.2$\pm$1.2 & 52.6$\pm$1.6 & 53.5$\pm$1.4 & 69.1$\pm$1.1 & 63.5$\pm$1.5 & 74.1$\pm$1.5 & 61.7$\pm$1.2 & 64.5$\pm$0.9 & 70.7$\pm$0.5 & 79.2$\pm$0.7 & 72.9$\pm$0.7 & 75.8$\pm$1.6 & 66.1 \\
    DANN (ICML15)~\cite{ganin2014unsupervised}  & 54.6$\pm$0.7 & 49.7$\pm$1.6 & 51.9$\pm$1.4 & 69.5$\pm$1.1 & 63.5$\pm$1.0 & 72.9$\pm$0.8 & 61.9$\pm$1.2 & 63.3$\pm$1.0 & 71.3$\pm$1.0 & 80.2$\pm$0.8 & 71.7$\pm$0.4 & 74.2$\pm$0.4 & 65.4 \\
    OSBP (ECCV18)~\cite{saito2018open} & 56.7$\pm$1.9 & 51.5$\pm$2.1 & 49.2$\pm$2.4 & 67.5$\pm$1.5 & 65.5$\pm$1.5 & 74.0$\pm$1.5 & 62.5$\pm$2.0 & 64.8$\pm$1.1 & 69.3$\pm$1.1 & 80.6$\pm$0.9 & 74.7$\pm$2.2 & 71.5$\pm$1.9 & 65.7 \\
    OpenMax (CVPR16)~\cite{bendale2016towards} & 56.5$\pm$0.4 & 52.9$\pm$0.7 & 53.7$\pm$0.4 & 69.1$\pm$0.3 & 64.8$\pm$0.4 & 74.5$\pm$0.6 & 64.1$\pm$0.9 & 64.0$\pm$0.8 & 71.2$\pm$0.8 & 80.3$\pm$0.8 & 73.0$\pm$0.5 & 76.9$\pm$0.3 & 66.7 \\
    STA (CVPR19)~\cite{liu2019separate} & 58.1$\pm$0.6 & 53.1$\pm$0.9 & 54.4$\pm$1.0 & 71.6$\pm$1.2 & 69.3$\pm$1.0 & 81.9$\pm$0.5 & 63.4$\pm$0.5 & 65.2$\pm$0.8 & 74.9$\pm$1.0 &\tblue{85.0$\pm$0.2} &\tblue{75.8$\pm$0.4} &\tblue{80.8$\pm$0.3} & 69.5 \\
    UAN (CVPR19)~\cite{Kaichao2019Universal}  &\tblue{59.4$\pm$0.2}  & \tblue{55.8$\pm$0.2}   & \tblue{62.4$\pm$0.3}   & \tblue{76.5$\pm$0.4}    &\tblue{70.1$\pm$0.4}  &\tblue{82.7$\pm$0.6}   &\tblue{65.8$\pm$0.7}   &\tblue{67.4$\pm$0.6}   &\tblue{75.0$\pm$0.8}   & 83.0$\pm$0.6   & 74.9$\pm$1.4  & 78.1$\pm$0.6  &\tblue{70.9} \\
    \hline
    \hline
    Ours  & \tred{63.7$\pm$0.3} & \tred{58.4$\pm$0.3} & \tred{64.4$\pm$0.1} & \tred{80.6$\pm$0.5} & \tred{74.2$\pm$0.4} & \tred{83.3$\pm$0.2} & \tred{68.4$\pm$0.3} & \tred{71.1$\pm$0.3} & \tred{78.0$\pm$0.4} & \tred{86.0$\pm$0.2} & \tred{79.5$\pm$0.2} & \tred{82.7$\pm$0.4} & \tred{74.2} \\
    \hline
    \end{tabular}%
\end{adjustbox}

  \label{tab:office-home}%
\end{table*}%

\begin{table*}[!t]
    \caption{Classification accuracy OS(\%)/Unk(\%) of OSDA on PACS (AlexNet). 
    }
    \begin{adjustbox}{width=2.0\columnwidth,center}
    \begin{tabular}{cccccccccccccc}
    \hline
    Method                               & Ar-Ph    & Ca-Ph & Sk-Ph & Ph-Ar & Ca-Ar & Sk-Ar & Ph-Ca & Ar-Ca & Sk-Ca & Ph-Sk & Ar-Sk & Ca-Sk & Avg.  \\
    \hline
    \hline
    AlexNet (NIPS2012)~\cite{krizhevsky2012imagenet} & 65.2/0.10 & 59.0/0.20 & 49.2/0.00 & 51.3/0.00 & 45.8/0.20 & 44.2/0.10 & 51.8/0.20 & 52.5/0.00 & 45.9/0.00 & 39.9/0.00 & 36.5/0.20 & 48.3/0.00 & 49.1/0.08 \\
    OSBP (ECCV18)~\cite{saito2018open}    & 62.7/\tblue{44.4} & 53.3/28.1 & 52.3/\tblue{57.1} & 50.8/41.8 & 42.5/14.6 & 40.0/\tblue{31.9} & 50.7\tred{/38.7} & 51.4/\tblue{37.6} & \tblue{57.1}/\tblue{38.5} & 45.7/16.3 & 41.4/30.3 & 52.3/26.5 & 50.0/33.8 \\
    STA (CVPR19)~\cite{liu2019separate} & 73.1/35.9 & \tblue{61.9}/\tblue{63.1} & 56.0/\tred{65.9} & 51.5/\tred{50.3} & \tblue{49.3}/\tblue{48.7} & 37.1/\tred{68.6} & 52.1/1.10 & 51.0/15.0& 55.5/4.20 & 48.5/\tred{78.5} & 42.9/\tblue{35.1} & 53.8/\tblue{50.5} & 52.7/\tblue{43.1} \\
    UAN (CVPR19)~\cite{Kaichao2019Universal} & \tblue{74.1}/23.7 & 60.9/14.1 & \tblue{62.9}/20.9 & \tred{56.5}/14.5 & \tred{49.9}/3.70 & \tblue{44.3}/19.1 & \tblue{54.8}/8.00 &\tblue{57.3}/4.50 & 48.2/21.2 & \tred{56.6/}6.20 & \tblue{52.2}/8.10 & \tblue{54.2}/3.80 & \tblue{56.0}/12.3 \\
    \hline
    \hline
    Ours & \tred{83.1}/\tred{63.9} & \tred{72.4}/\tred{71.5} & \tred{69.6}/27.0 & \tblue{55.3}/\tblue{45.4} & 48.6/\tred{53.7} & \tred{44.8}/25.3 & \tred{67.8}/\tblue{36.6} & \tred{59.8}/\tred{49.9} & \tred{63.2}/\tred{60.5} & \tblue{56.2}/\tblue{72.4} & \tred{55.8}/\tred{50.7} & \tred{63.1}/\tred{70.0} & \tred{61.6}/\tred{52.2} \\
    \hline
    \end{tabular}%
    
\end{adjustbox}

  \label{tab:pacs-alexnet}%
\end{table*}%

\begin{table*}[!t]
      \caption{Classification accuracy  OS(\%)/Unk(\%) of OSDA on PACS (ResNet-$50$).
      }
      \begin{adjustbox}{width=2.0\columnwidth,center}
    \begin{tabular}{cccccccccccccc}
    \hline
    Method & Ar-Ph & Ca-Ph & Sk-Ph & Ph-Ar & Ca-Ar & Sk-Ar & Ph-Ca & Ar-Ca & Sk-Ca & Ph-Sk & Ar-Sk & Ca-Sk & Avg.  \\
    \hline
    \hline
    ResNet50 (CVPR16)~\cite{he2016deep}      & 73.5/0.00 & 65.4/0.30 & 67.9/0.40 & 63.3/0.00 & 56.7/0.30 & 48.5/0.20 & 51.5/0.20 & 48.2/0.50 & 50.2/18.4 & 39.1/0.10 & 37.9/0.10 & 45.5/0.20 & 54.0/1.73 \\
    OSBP (ECCV18)~\cite{saito2018open} & 86.4/55.8 & 64.4/38.0 & 69.9/\tred{77.4} & 71.9/37.7 & 57.5/42.5 & 58.1/58.4 & 58.5/36.3 & 61.8/27.0 & 62.9/23.0 & 52.1/41.6 & 49.2/43.2 & 50.9/22.1 & 62.0/41.9 \\
    STA (CVPR19)~\cite{liu2019separate} & \tblue{87.9}/\tblue{59.1} & 70.0/\tblue{39.7} & \tblue{77.0}/47.9 & 67.8/\tblue{71.1} & 62.6/\tblue{47.6} & 52.1/\tred{75.5} & 56.5/\tblue{45.7} & 57.3/\tblue{43.5} & \tblue{64.2}/\tblue{44.7} & 55.3/\tblue{64.3} & \tblue{55.0}/\tblue{52.6} & \tblue{58.1}/\tblue{35.9} & 63.7/\tblue{52.6} \\
    UAN (CVPR19)~\cite{Kaichao2019Universal} & 84.8/43.4 & \tblue{72.6}/4.84 & 73.3/23.8 & \tblue{68.8}/7.60 & \tblue{64.2}/5.40 & \tred{67.5}/38.4 & \tred{69.1}/19.0 & \tred{64.3}/19.3 & 61.4/1.70 & \tblue{55.7}/17.9 & 40.4/16.6 & 53.9/4.70 & \tblue{66.3}/16.9 \\
    \hline
    \hline
    Ours  & \tred{91.2}/\tred{71.7} & \tred{83.0}/\tred{54.2} & \tred{83.5}/\tblue{61.9} & \tred{75.8}/\tred{71.9} & \tred{69.1}/\tred{56.2} & \tblue{67.4}/\tblue{60.2} & \tblue{60.9}/\tred{57.9} & \tblue{62.6}/\tred{59.6} & \tred{72.3}/\tred{45.9} & \tred{62.8}/\tred{69.7} & \tred{60.7}/\tred{60.1} & \tred{65.7}/\tred{65.1} & \tred{71.3}/\tred{61.2} \\
    \hline
    \end{tabular}%
\end{adjustbox}

  \label{tab:paccs-resnet}%
\end{table*}%

\subsection{Evaluation on Office-Home dataset}

\subsubsection{Setting} 
Office-Home contains about $15,500$ images from 4 different domains: Artistic (\textbf{Ar}), Clipart (\textbf{Cl}), Product (\textbf{Pr}) and Real-World (\textbf{Rw}), with each domain containing images from 65 object classes. 
Gathering an idea from earlier works~\cite{liu2019separate} we denote the first $25$ classes (in alphabetic order) as classes shared by the source and target domains whereas classes $26$-$65$ belong to the unknown class. We construct open set domain adaptation tasks between two domains in both directions, forming $12$ tasks where domain discrepancy is larger than Office-$31$. Domain adversarial network used is the same as the one used in last evaluation~\cite{ganin2014unsupervised}, with ResNet-$50$ as the backbone~\cite{liu2019separate}. This time, only one evaluation metric is employed due to space limitations. ~\textbf{OS}: normalized accuracy for all the classes including the unknown as one class; The optimization parameters and execution statistics are identical to the previous evaluation except that we set $\alpha$=$0.5$ and $\beta$= $0.3$.


\subsubsection{Results}
From results in Table~\ref{tab:office-home}, we can see that: 
(i) On certain tasks, the method having ResNet backbone, outperforms OSDA method due to the adverse effects of unknown classes on domain adaptation. Significant gaps cross domains and label spaces add to the problem of negative transfer brought forth by unknown classes, thus collapsing performance.
(ii) Our method exceeds the performance of existing methods by significant margins on all tasks. 
(iii) Overall, our method ranks the highest, improving $8.9$\% (OS) on ResNet-$50$, and $3.3$\% (OS) on prior state-of-the-art UAN~\cite{Kaichao2019Universal}.

\begin{table*}[!t]
    \caption{
    Ablative study of our model showing classification accuracy OS(\%)/Unk(\%) on PACS (ResNet-$50$). 
    }
    \begin{adjustbox}{width=2.0\columnwidth,center}
    \begin{tabular}{cccccccccccccc}
    \hline
    Method & Ar-Ph & Ca-Ph & Sk-Ph & Ph-Ar & Ca-Ar & Sk-Ar & Ph-Ca & Ar-Ca & Sk-Ca & Ph-Sk & Ar-Sk & Ca-Sk & Avg.  \\
    \hline
    \hline
    MTS w/o w & 83.0/43.1 & 75.3/39.3 & 76.0/38.2 & 73.2/\tred{72.3} & 66.2/46.7 & 54.9/0.00 & 59.6/0.20 & 63.3/\tblue{48.9} & 63.5/41.0 & 44.4/0.00 & 49.8/50.1 & 58.3/55.0 & 64.0/36.2 \\
    MTS w/o mutual & \tblue{85.3}/\tblue{63.6} & 77.7/\tblue{63.0} & 77.3/52.7 & 72.1/64.3 & 67.5/39.0 & 61.6/55.3 & 59.1/0.00 & \tblue{63.7}/43.0 & 64.3/44.5 & 48.6/0.00 & 53.8/\tblue{66.7} & 60.8/\tred{68.3} & 66.0/46.7  \\
    MTS w/o ds & 85.2/49.5 & 76.4/50.0 & 76.7/56.6 & 71.4/59.1 & 67.7/49.1 & 57.6/44.1 & 59.8/0.10 & 62.6/47.3 & 65.2/44.8 & 50.1/0.00 & 44.0/0.00  & 56.6/61.1 & 66.4/38.5  \\
    MTS w/o mse & 85.1/47.9 & 77.8/60.1 & 78.5/\tred{67.1} & 73.6/68.8 & 68.7/54.9 & 62.7/35.7 & 59.7/\tred{63.6} & 62.5/39.7 & 65.3/\tred{50.0} & 50.9/17.9 & 52.6/54.1 & \tblue{61.9}/43.9 & 66.6/50.3 \\
    MTS w/o s & 85.2/44.1 & \tred{84.8}/\tred{67.4} & \tblue{80.0}/46.4 & \tblue{75.5}/60.2 & \tred{70.2}/\tred{58.5} & \tred{69.7}/\tblue{56.8} & \tred{61.0}/47.8 & \tred{67.5}/47.1 & \tblue{69.5}/40.7 & \tblue{61.6}/\tblue{58.0} & \tblue{60.4}/\tred{70.6} & 60.6/\tblue{65.4} & \tblue{70.5}/\tblue{55.3}\\
    \hline
    \hline
    MTS  & \tred{91.2}/\tred{71.7} & \tblue{83.0}/54.2 & \tred{83.5}/\tblue{61.9} & \tred{75.8}/\tblue{71.9} & \tblue{69.1}/\tblue{56.2} & \tblue{67.4}/\tred{60.2} & \tblue{60.9}/\tblue{57.9} & 62.6/\tred{59.6} & \tred{72.3}/\tblue{45.9} & \tred{62.8}/\tred{69.7} & \tred{60.7}/60.1 & \tred{65.7}/65.1 & \tred{71.3}/\tred{61.2} \\
    \hline
    \end{tabular}%
\end{adjustbox}
  \label{tab:ablation}%
\end{table*}%

\begin{figure*}[!t] 
  \centering  

  \subfigure[]{
    \label{fig:vis:1a}
    \includegraphics[width=0.45\linewidth]{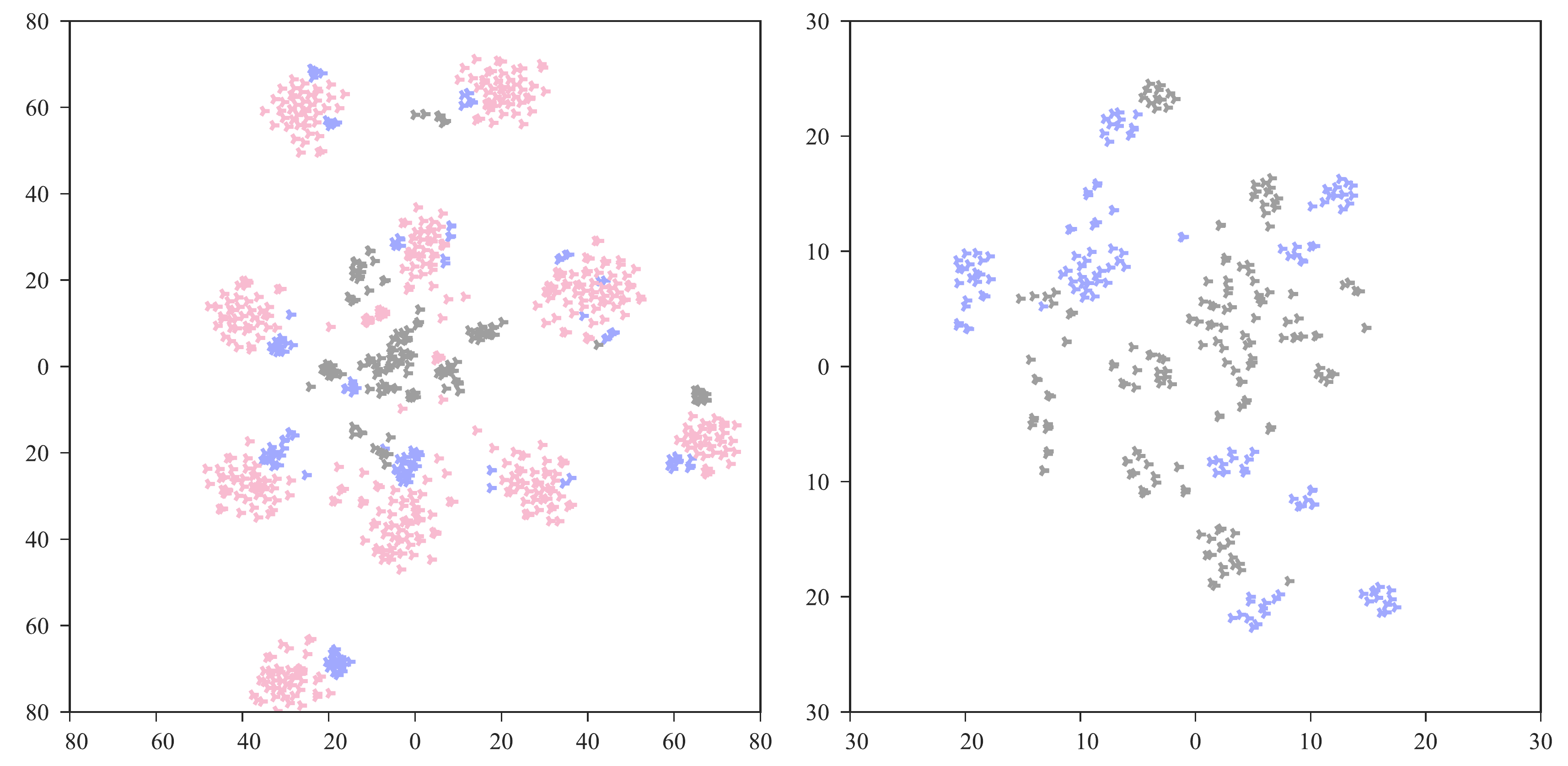}}
  \quad 
  \subfigure[]{
    \label{fig:vis:1b}
    \includegraphics[width=0.45\linewidth]{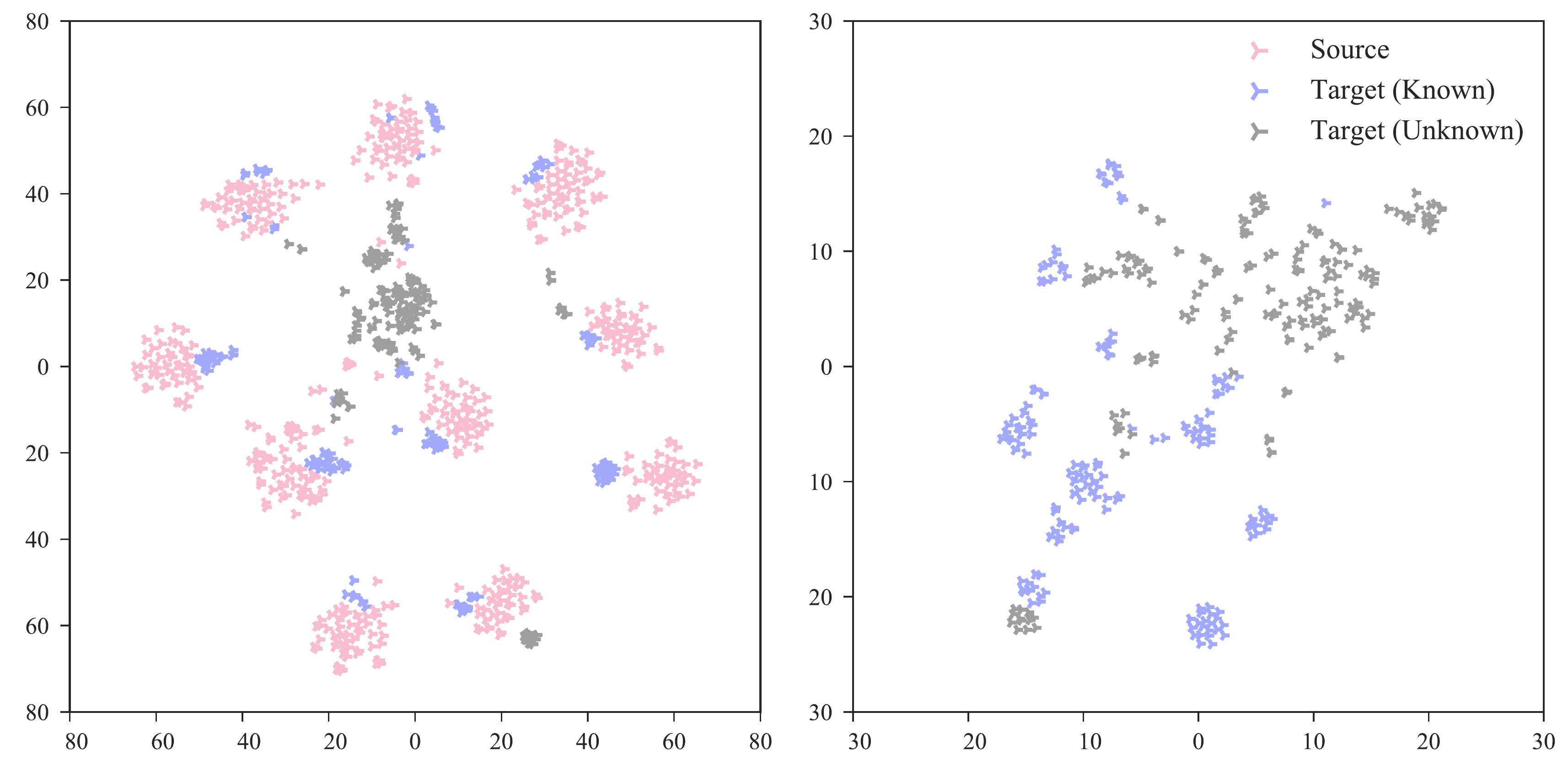}}  
  \caption{Visualization of the features extracted by (a) ResNet$50$ and (b) Our method, on task A $\rightarrow$ D using t-SNE embeddings, respectively. Pink, blue, and grey points refer to source features, target features of known classes and target features of unknown classes, respectively.~\emph{Best viewed in color and zoomed in.}}
  \label{fig:vis}

\end{figure*}

\begin{table}[!t]
  \caption{ Average Classification Accuracy (\%) of OSDA tasks on Office-$31$ (ResNet$50$, pre-trained) and Office-$31$ (ResNet-$18$, trained from scratch).
  }
\begin{adjustbox}{width=0.9\columnwidth,center}
    \begin{tabular}{ccccccc}
    \hline
    \multirow{2}*{Method} & \multicolumn{3}{c}{ResNet50} & \multicolumn{3}{c}{ResNet18} \\
    \cline{2-7}          & OS    & OS*   & Unk   & OS    & OS*   & Unk \\
    \hline
    \hline
    ResNet18 (CVPR16)~\cite{he2016deep} & 84.2  & 84.4  & \tblue{82.2}  & 33.9    & 36.3  & 0.00 \\
    OSBP (ECCV18)~\cite{saito2018open}   & 90.8  & 91.3  & \tred{85.8} & 36.5  & 35.3  & \tblue{48.5} \\
    STA (CVPR19)~\cite{liu2019separate}   & \tblue{92.9}  & 94.1  & 80.9  & 36.4  & 35.9  & 41.4 \\
    UAN (CVPR19)~\cite{Kaichao2019Universal}    & 89.3  & \tred{97.1} & 11.3  & \tblue{36.6}  & \tblue{39.9}  & 3.60\\
    \hline
    \hline
    Ours  & \tred{93.8} & \tblue{96.4}  & 67.8  & \tred{43.9 } & \tred{42.9 } & \tred{53.9}\\
    \hline
    \end{tabular}%
\end{adjustbox}

  \label{tab:office31-2}%
\end{table}%

\subsection{Evaluation on PACS dataset}

\subsubsection{Re-purposing for Open-set Setting} 
PACS was originally proposed as a domain generalization benchmark~\cite{Li2017dg}, which we have re-purposed for a more ambitious challenge of open set domain adaptation. It contains around $9,991$ images from $4$ different domains.
For this evaluation we have chosen the first $3$ classes alphabetically as classes shared by the source and target domains, while the next $4$-$7$ classes belong to the unknown class. We construct OSDA tasks between two domains in both directions, forming $12$ tasks. Besides having a higher practical relevance, re-purposed PACS has greater domain shift than other existing benchmarks as shown in Figure~\ref{fig:OSDA}, thus posing to be a challenging dataset. 
This time AlexNet and ResNet-$50$ are used as backbone with the same domain adversarial network~\cite{ganin2014unsupervised}.  
The following two evaluation metrics have been employed here: 
~\textbf{OS}: normalized accuracy for all the classes including the unknown as one class;
~\textbf{Unk}: the accuracy of unknown samples. Optimization parameters and execution statistics, are identical to those used during evaluation on Office-$31$ (\S 4.2), except that we set $\alpha$ = $0.6$ and $\beta$=$0.3$ for AlexNet and $\alpha$ = $0.8$ and $\beta$=$0.6$ for ResNet-$50$. 
Standard deviation enumerated in Tables~\ref{tab:pacs-alexnet} and~\ref{tab:paccs-resnet} have been omitted due to space limitations. \\

\subsubsection{Results}
Results in Tables~\ref{tab:pacs-alexnet} and~\ref{tab:paccs-resnet}, show that:
(i) Our method surpasses existing methods significantly in performance on most tasks. 
(ii) With an AlexNet backbone, our method secures an improvement of $5.6$\% (OS) on prior state-of-the-art UAN and a $9.1$\% (Unk) on prior state-of-the-art STA.
(iii) Using a ResNet-$50$ backbone, our method improves by $5.0$\% (OS) on prior state-of-the-art UAN and by $8.6$\% (Unk) on prior state-of-the-art STA.

\subsection{Analysis} 

\subsubsection{Ablation Study} 
We conducted an ablation study using PACS-ResNet50 to justify the contribution of each component of our proposed method, as shown in Table~\ref{tab:ablation}.
(\textbf{1}) MTS outperforms \textbf{MTS w/o w}, the variant without weighting target samples in domain adversarial adaptation. This indicates that aligning samples of unknown classes with source samples leads to negative transfer and performance degradation. As a result it confirms the necessity of weights that we had adopted to separate samples of known and unknown classes.
(\textbf{2}) MTS without mutual learning (\textbf{MTS w/o mutual}) performs lower than our method, indicating that mutual learning increases the separation between samples of unknown and known classes, which is beneficial for domain adaptation without unknown samples. 
(\textbf{3}) Inferior performance of MTS without domain-separate loss (\textbf{MTS w/o ds}) compared to our method, indicates, that this loss can indeed repel outlier target samples from source and known target samples during domain adaptation.
(\textbf{4}) Substituting European distance with KL divergence, \textbf{MTS w/o mse} scores lesser than our method. This shows that European distance provides a better measurement of the similarity between two distributions in this situation.
(\textbf{5}) Compared to \textbf{MTS w/o s} that shares parameters between two CNN networks in both components of our model, MTS secures considerably higher results. This signifies that non-shared parameters between two CNN networks can generate stronger feature representation, thus being better suited to mutual learning.
From this ablative study it can be confirmed that every component of our method has a contribution towards the final performance. 

\subsubsection{Feature Visualization} 
We visualize the last-layer features extracted by ResNet$50$ and our proposed method MTS on task Amazon $\rightarrow$ DSLR in Figure~\ref{fig:vis:1a} and Figure~\ref{fig:vis:1b}.
As shown in Figure~\ref{fig:vis:1a}, features of several known classes are close or even mixed together with the unknown classes. This shows that ResNet fails to discriminate between them during training. Conversely, Figure~\ref{fig:vis:1b} shows that MTS is capable of aligning target features of known classes, to source features in the target domain, with better accuracy, while distancing features of unknown classes far apart. 
This clearly establishes superiority of our method over others.

\subsection{Evaluation on Office-$31$ dataset in a fairer sense}

\subsubsection{Setting}  
From the results in Table~\ref{tab:office31-2}, we observe that the baseline method has the potential to recognize unknown samples (82.2\% accuracy on unknown class) when we use the pre-trained ResNet-$50$ for fine-tuning. One of the probable reasons for this might be that the model is already aware of the unknown samples before training as it was pre-trained.
Therefore, enforcing a fairer comparison, we employ ResNet-$18$ as a backbone, and train from scratch. Momentum SGD is selected as the optimizer, with a learning rate of $0.1$ set initially and multiplied by $0.1$ at $150$th and $225$th epochs, successively. We train our model for 300 epochs, with a momentum of $0.9$, and weight decay of $5\times10^{-4}$. We set $\alpha$ = $0.6$ and $\beta$=$0.2$. Rest of the settings are consistent with those used during evaluation on Office 31 (\S~\ref{subsection:office31}).

\subsubsection{Results} 
Results in Table~\ref{tab:office31-2} show that on using a ResNet-$18$ backbone:
(i) The baseline method is unable to recognize unknown samples.
(ii) Our method outperforms existing methods by broad margins on most tasks with different evaluation metrics. 
(iii) Overall, our method achieves the highest performance score improving: $10.9$\% on (OS), $6.6$\% on (OS*), and  $53.9$\% on (Unk) with ResNet-$18$;
$7.3$\% on (OS), $3.0$\% on (OS*) with prior state-of-the-art UAN;
and $5.4$\% on (OS) with prior state-of-the-art STA.
(iv) Even without using a pre-trained model our method delivers a better performance. This further proves the robustness and superiority of our method over other existing state-of-the-arts.



\section{Conclusion}

In this paper, we studied the problem of open set domain adaptation (OSDA) paying particular attention towards tackling the domain gap. We first observed a significant performance drop from state-of-the-art methods, under the presence of larger domain gaps. We attributed this to the inability of existing models to cultivate the mutual relationship between unknown sample classification and domain distribution matching. We have therefore proposed a mutual learning framework, where two networks are specifically designed to motivate positive information exchange. We show via experiments that our method outperforms state-of-the-arts on representative OSDA benchmarks,~\emph{i.e.}, Office-$31$, Office-Home, and especially when the domain gap is large, on the PCAS dataset which we re-purposed for OSDA for the first time.
Our in-depth ablative study further validates the contribution of every component in our model towards its superior performance.


%





\ifCLASSOPTIONcaptionsoff
  \newpage
\fi



%

{\small
\bibliographystyle{ieee}
\bibliography{main}
}




%







\end{document}